\documentclass[sigplan,nonacm]{acmart}

\usepackage{booktabs}
\usepackage{listings}
\usepackage{tikz}
\usetikzlibrary{arrows.meta,positioning,shapes.geometric,fit,calc,backgrounds}

\renewcommand\footnotetextcopyrightpermission[1]{}
\acmConference[Agent Skills '26]{Agent Skills '26}{May 26, 2026}{San Jose, CA}
\acmYear{2026}
\copyrightyear{2026}
\setcopyright{none}
\title[How Much Heavy Lifting Can an Agent Harness Do?]{How Much Heavy Lifting Can an Agent Harness Do?: Measuring the LLM's Residual Role in a Planning Agent}

\author{Sungwoo Jung}
\affiliation{%
  \institution{Independent Researcher}
  \country{}
}
\email{sigran0@gmail.com}

\author{Seonil Son}
\affiliation{%
  \institution{RLWRLD.AI}
  \city{Seoul}
  \country{South Korea}
}
\email{simon.son@rlwrld.ai}

\begin{document}

\begin{abstract}
Agent harnesses---the stateful programs that wrap a language model and decide what it sees at each step---are now known to change end-to-end performance on a fixed model by as much as six times. That raises a question asked less often than it should be: how much of an agent's competence does the harness itself already carry, and how much genuinely still needs the LLM? We externalize a planning harness for noisy Collaborative Battleship into four progressively richer layers---posterior belief tracking, declarative planning, symbolic reflection, and an LLM-backed revision gate---under a common runtime, taking \emph{win rate} as the primary metric and \emph{F1} as secondary, and pre-specifying \emph{heavy lifting} as the single largest positive marginal to the primary metric. Across 54 games, declarative planning carries the heavy lifting ($+24.1$pp win rate over a belief-only harness, zero LLM calls); symbolic reflection is mechanistically real but calibration-sensitive, with signed board-level effects up to $\pm0.140$ F1 that cancel on aggregate; and LLM-backed revision activates on only $4.3\%$ of turns with a bounded, non-monotonic effect. The contribution is methodological: once harness layers are made externally measurable, the LLM's role can be quantified as residual rather than assumed central.
\end{abstract}

\keywords{agent harness, harness engineering, harness decomposition, structure-efficacy relationship, layer-wise ablation, LLM-based planning, residual LLM intervention, declarative runtime}

\maketitle

\section{Introduction}

Harness engineering has become a load-bearing concept for LLM agents in 2025--2026. Recent industry and research reports show that the wrapper around a fixed language model---what gets stored, retrieved, and presented at each step---can change end-to-end performance by as much as six times on the same benchmark~\cite{boluk2026harnessproblem,bockeler2026harnessengineering,openai2026harnessengineering,young2025harness}. A harness, in this sense, is a stateful program that wraps a language model and determines what context the model sees at each step. Production systems such as Claude Code~\cite{anthropic2025claudecode}, Gemini CLI, Codex, and OpenClaw lean heavily on harness structure---including shared procedural artifacts such as the \texttt{SKILL.md} spec~\cite{agentskills}---to obtain reliable behavior from a general-purpose language model.

If the harness can change performance by six times, however, a question follows that is asked less often than it should be: \emph{how much of an agent's competence does the harness itself already carry, and how much genuinely still needs the LLM?} Most current harnesses bundle belief tracking, action selection, reflection, and revision inside a single LLM-orchestrated loop~\cite{yao2023react,shinn2023reflexion}, which makes it hard to tell which layer is doing the heavy lifting.

Recent work on evolving agent systems---Agentic Context Engineering~\cite{zhang2025ace}, Meta Context Engineering~\cite{ye2026meta}, Automated Design of Agentic Systems~\cite{hu2025automated}, and minimal-harness systems such as Terminus-KIRA~\cite{terminuskira2026} on Terminal-Bench~\cite{merrill2026terminal}---\emph{searches over} entire harnesses. It does not typically \emph{decompose a single} harness into internally measurable layers, which is the gap this paper addresses.

We address this gap in one narrow, controlled setting. We take a planning harness and externalize it into four progressively richer layers: (L1) posterior belief tracking; (L2) declarative planning over the posterior through hypothetical transition evaluation and question timing; (L3) symbolic reflection via a confidence-gated, LLM-free revision mechanism; and (L4) LLM-backed revision activated only when the confidence gate opens. The planning domain is noisy Collaborative Battleship~\cite{grand2025battleship}, a controlled mini-lab in which partial observability, belief update, budgeted questioning, and uncertainty-aware action selection coexist with turn-level consequences. Each layer is lifted into an inspectable declarative runtime so that its marginal contribution to the agent's performance, and the LLM's eventual invocation rate, can be observed rather than assumed.

\paragraph{Pre-specified reporting criterion.}
Because layer contributions can diverge across metrics, we fix the reporting protocol before presenting results. We take \emph{win rate} as the primary metric (it reflects whether the agent completes a game, the terminal success criterion of the domain) and \emph{F1} as a secondary metric (it reflects local targeting precision conditional on a fixed shot budget). We pre-specify that a layer qualifies as \emph{heavy-lifting} iff it contributes the single largest positive marginal to the primary metric among the measured layers; under this criterion at most one layer can be heavy-lifting. All per-layer comparisons in Section~4 are evaluated against this rule, and cases where F1 and win-rate ranks disagree are reported explicitly rather than resolved silently.

Across 54 games, the planning layer carries most of the heavy lifting under the above criterion (+24.1pp win rate over a belief-only harness without a single LLM call); symbolic reflection is mechanistically real yet not net-positive on aggregate; and LLM-backed revision activates on only 4.3\% of turns and yields a bounded, non-monotonic residual effect. The resulting picture is harness-first and LLM-residual. We make three contributions:

\begin{enumerate}
  \item \textbf{Framework.} A four-layer decomposition protocol for a planning harness that isolates belief, planning, symbolic reflection, and LLM-backed revision as independently measurable runtime layers.
  \item \textbf{Evidence.} Per-layer marginal contributions on 54 games of noisy Collaborative Battleship, with the LLM invocation rate emerging as a \emph{dependent variable} of the confidence gate rather than a configured quota.
  \item \textbf{Design bias.} A direct answer to ``how much heavy lifting can an agent harness do?'' in this setting: most of the agent's competence is carried by the first three layers of the harness, and the LLM's genuine responsibility is sparse, gated, and bounded.
\end{enumerate}

\section{What an Agent Harness Actually Does}

\paragraph{Harness as load-bearing substrate.}
A harness is a stateful program that wraps a language model and determines what context the model sees at each step~\cite{bockeler2026harnessengineering,openai2026harnessengineering}. Because the harness can change benchmark performance on a fixed model by as much as six times~\cite{boluk2026harnessproblem}, it is as load-bearing as model weights for many production use cases. The Agent Skills ecosystem~\cite{agentskills} makes some procedural structure textual (\texttt{SKILL.md} is adopted by 30+ platforms), but most of the structural work in production harnesses is still load-bearing on the LLM inside the loop.

\paragraph{What current harnesses bundle.}
Production planning harnesses routinely mix belief tracking, planning, reflection, and revision inside a single LLM-orchestrated loop. An LLM is asked, at each step, to decide what to believe, what to do, what went wrong, and how to repair. When the whole loop is driven by prompts, it is hard to diagnose which internal component of the harness carried---or broke---the agent. \emph{This paper studies one narrow empirical question in this regime: how much heavy lifting can the harness itself do before the LLM is genuinely needed?}

\paragraph{Decomposition vs.\ search.}
The recent harness literature splits along two axes. One axis is \emph{search}: Agentic Context Engineering~\cite{zhang2025ace}, Meta Context Engineering~\cite{ye2026meta}, Automated Design of Agentic Systems~\cite{hu2025automated}, and Terminus-KIRA~\cite{terminuskira2026} evolve or design entire harnesses, asking \emph{which} macro-harness performs best on a benchmark such as Terminal-Bench~\cite{merrill2026terminal}. The orthogonal axis, addressed here, is \emph{decomposition}: holding a single harness fixed, lifting each internal layer into a separately addressable runtime object, and asking \emph{what each layer contributes independent of the LLM}. Search and decomposition are complementary---search yields the right macro-structure; decomposition yields per-layer attribution---but they require different instrumentation. Search ranks harnesses by end-to-end pass rate; decomposition requires the harness to be lifted into objects whose preconditions and effects are externally observable. The unit of measurement, not the unit of optimization, is what changes.

\paragraph{Four layers, four prior-work analogs.}
The four layers we ablate are not invented from scratch; each isolates a line of prior work that current harnesses bundle into a single LLM-orchestrated loop (Table~\ref{tab:layer-prior}). L1 isolates the Bayesian belief layer of classical experimental design and the original Battleship harness~\cite{grand2025battleship}. L2 isolates the program-guided / world-model planning layer of LLM+P~\cite{liu2023llmp} and WorldCoder~\cite{tang2024worldcoder}, with the LLM lifted out so the planning layer's marginal can be measured against a fixed belief backend. L3 isolates the meta-cognitive / self-reflective layer of Reflexion~\cite{shinn2023reflexion} and MIDCA~\cite{cox2016midca}, replacing the LM-emitted reflection with a confidence-gated symbolic mechanism so the gate's effect is measurable independent of LM output quality. L4 reattaches LM intervention as a \emph{conditional residual} on the same gate---in the spirit of ReAct~\cite{yao2023react}, but bounded to turns where the symbolic substrate has already declared itself uncertain. Each layer can therefore be ablated against its prior-work analog without changing the underlying mechanism.

\begin{table}[t]
\centering
\scriptsize
\caption{The four harness layers and the prior-work line each isolates. Decomposition is what makes each $\Delta_i$ measurable against a fixed substrate rather than against a different bundle.}
\label{tab:layer-prior}
\begin{tabular}{@{}p{0.18\columnwidth}p{0.36\columnwidth}p{0.36\columnwidth}@{}}
\toprule
Layer & What it isolates & Closest prior-work analog \\
\midrule
L1. Belief             & Posterior over hidden world state & Bayesian experimental design; \cite{grand2025battleship} belief backend \\
L2. Planning           & World-model rollout for action and question scoring & LLM+P~\cite{liu2023llmp}, WorldCoder~\cite{tang2024worldcoder} \\
L3. Symbolic reflection & Confidence-gated, LLM-free in-episode revision & Reflexion~\cite{shinn2023reflexion}, MIDCA~\cite{cox2016midca} \\
L4. LLM-backed revision & Conditional LM intervention on the same gate & ReAct~\cite{yao2023react} \\
\bottomrule
\end{tabular}
\end{table}

\paragraph{Why bundling obscures attribution.}
Let $\mathcal{H} = (B, P, R, V)$ denote a planning harness decomposed into belief, planning, reflection, and revision components, and let $\Phi(\mathcal{H})$ be end-to-end performance on a fixed metric. The per-layer marginal contribution of component $i$ is $\Delta_i := \Phi(\mathcal{H}) - \Phi(\mathcal{H}_{-i})$, where $\mathcal{H}_{-i}$ is the harness with component $i$ removed or replaced by a neutral substitute. Current harnesses fall into one of two regimes:
\begin{itemize}\itemsep 0.1em
  \item \emph{Bundled.} A single LLM call $\ell(\cdot)$ jointly emits belief updates, action proposals, reflective judgments, and revisions, so
    $$\Phi(\mathcal{H}_{\text{bundle}}) = \Phi\bigl(\ell(s_t;\,\pi)\bigr),$$
    where $\pi$ is a prompt pattern. The individual $\Delta_i$ are not identifiable from end-to-end traces alone, because removing any one role from the prompt also changes the others through their shared generation distribution.
  \item \emph{Layered.} Each component is a separately callable runtime object with its own inputs, outputs, and (optionally) an LLM dependency, so
    $$\Phi(\mathcal{H}_{\text{layered}}) = \Phi\bigl(B(s_t),\; P(\cdot),\; R(\cdot),\; V_{\text{det}}(\cdot) \oplus V_M(\cdot)\big|_{\text{gate}}\bigr),$$
    and each $\Delta_i$ is directly measurable by a single-component ablation. Attribution is well-posed iff each component is lifted out of the shared LLM context.
\end{itemize}
This paper operates in the layered regime and measures $\Delta_i$ for $i \in \{B, P, R, V\}$ on a shared Battleship backend.

\section{Externalizing the Planning Harness}

\paragraph{Preliminaries: state, signals, and gates.}
Let $s_t$ denote the game state at turn $t$ and $o_t$ the observation (a noisy hit/miss return or a question answer). The harness maintains a particle-approximated posterior $B_t$ over hidden ship placements (500 particles, Metropolis--Hastings backend), a per-turn predictive error $e_t^{\text{pred}}$ and calibration error $e_t^{\text{cal}}$, and their EMAs $\bar{e}_t^{\text{pred}}, \bar{e}_t^{\text{cal}}$ with coefficient $\alpha\in(0,1]$. We define the runtime-computed \emph{model confidence}
$$c_t \;:=\; 1 - \tfrac{1}{2}\bigl(\bar{e}_t^{\text{pred}} + \bar{e}_t^{\text{cal}}\bigr) \;\in\; [0,1],$$
and say the \emph{revision gate is open} at turn $t$ iff (i) $c_t < \tau$ for the current threshold $\tau$, (ii) the low-confidence streak has length $\geq k_{\text{streak}}$, (iii) the revision cooldown counter is zero, and (iv) a \texttt{sim.next} counterfactual preview yields $\Delta\Phi \geq \delta_{\min}$ against the current policy. All four conditions are runtime-computable; none require an LLM. The layers differ only in (a) whether the gate's truth value is consulted and (b) who writes the revision patch when the gate opens (a symbolic preset vs.\ an LLM).

\paragraph{Battleship as a decomposition lab.}
We study this in noisy Collaborative Battleship~\cite{grand2025battleship}, where the agent maintains a posterior over hidden ship placements, chooses shots and information-gathering questions under a budget, and acts under noisy observations. Belief update, budgeted information gathering, uncertainty-aware action selection, and in-episode revision all appear at turn-level granularity in a compact domain. Battleship is not a canonical harness benchmark, and we do not use it as one. It serves here as a controlled mini-lab: a domain in which one harness layer can be varied at a time with interpretable turn-level consequences, ahead of replicating the decomposition in messier agent settings.

\begin{table}[t]
\centering
\scriptsize
\caption{Progressive externalization of a planning harness. Each row adds one layer that the harness carries without calling the LLM; the LLM enters only at L4, and its invocation rate is measured rather than pre-budgeted.}
\label{tab:stack-map}
\begin{tabular}{@{}p{0.24\columnwidth}p{0.46\columnwidth}p{0.17\columnwidth}@{}}
\toprule
Harness layer & Role added on top of the previous layer & LLM? \\
\midrule
L1. Belief-only          & Posterior belief tracking & No \\
L2. + Planning           & Hypothetical transition evaluation and question timing & No \\
L3. + Symbolic reflection & Confidence tracking and guarded revision actions & No \\
L4. + LLM-backed revision & Residual revision when the confidence gate opens & Conditional \\
\bottomrule
\end{tabular}
\end{table}

\begin{figure}[t]
\centering
\begin{tikzpicture}[
  every node/.style={font=\scriptsize},
  node distance=3mm,
  phase/.style={draw,rounded corners=2pt,minimum width=6.2cm,minimum height=8.5mm,align=center,inner sep=2.5pt},
  setup/.style={phase,fill=black!5,minimum width=4.0cm,minimum height=6mm},
  p1/.style={phase,fill=cyan!10,draw=cyan!55!black},
  p2/.style={phase,fill=violet!10,draw=violet!60!black},
  p3/.style={phase,fill=red!10,draw=red!55!black},
  p4/.style={phase,fill=orange!15,draw=orange!60!black},
  arr/.style={-{Latex[length=1.8mm,width=1.5mm]},semithick},
  lbl/.style={font=\tiny\itshape,text=black!60},
]
\node[setup] (S) {Setup: \texttt{initBoard}};
\node[p1,below=10mm of S] (P1) {\textbf{1. Observe \& evaluate}\\[0.5pt] world summary $\to$ decision context $\to$ 9 previews};
\node[p2,below=of P1] (P2) {\textbf{2. Decide \& act} (\texttt{sim.next}-evaluated)\\[0.5pt] \texttt{preferQuestion} $\oplus$ \texttt{shoot}};
\node[p3,below=of P2] (P3) {\textbf{3. Reflect}\\[0.5pt] prediction error $\to$ confidence EMA, streaks};
\node[p4,below=of P3] (P4) {\textbf{4. Revise} \,(L3: symbolic \;/\; L4: LLM-backed)\\[0.5pt] \texttt{shouldRevisePolicy} $\to$ apply preset / patch};
\node[setup,below=7mm of P4] (E) {\texttt{endGame(result)}};

\draw[arr] (S) -- (P1);
\draw[arr] (P1) -- (P2);
\draw[arr] (P2) -- (P3);
\draw[arr] (P3) -- (P4);
\draw[arr] (P4) -- node[right,lbl,pos=0.5]{\texttt{allShipsSunk} $\lor$ \texttt{shots}=0} (E);

\draw[arr,densely dashed,black!55]
  (P4.east) -- ++(5mm,0) |- node[pos=0.25,right,lbl]{next turn} (P1.east);

\begin{pgfonlayer}{background}
\node[draw,densely dashed,rounded corners=3pt,inner sep=2mm,
      fit=(P1)(P4)] (loopbox) {};
\end{pgfonlayer}
\node[anchor=south west,lbl] at ([xshift=1.5mm,yshift=-0.5mm]loopbox.north west)
  {Main loop: while \texttt{shotsRemaining>0} $\land$ \texttt{not allShipsSunk}};

\end{tikzpicture}
\caption{Planning harness main loop, externalized into declarative runtime events. Layers L1--L4 in Table~\ref{tab:stack-map} correspond to which phases are active: L1 uses only the belief-driven \texttt{shoot} in phase~2; L2 adds \texttt{sim.next}-based action selection with question timing; L3 adds the Reflect and Revise phases with symbolic presets; L4 replaces the Revise body with an LLM-backed patch when the confidence gate opens.}
\label{fig:loop}
\end{figure}
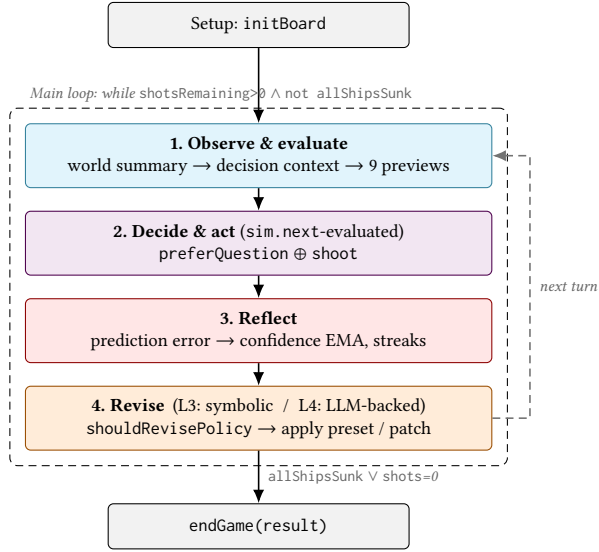

\paragraph{Declarative runtime as instrumentation.}
The four layers in Table~\ref{tab:stack-map} are lifted into a declarative runtime whose primitives are \emph{state}, \emph{computed properties}, \emph{guarded actions}, and \emph{patches}. State is a typed record updated only by patches. Computed properties (e.g., \texttt{modelConfidence}, \texttt{sustained}, \texttt{positivePreview} in Listing~\ref{lst:gate}) are pure functions over the state record, recomputed on every read and unable to have side effects. Actions declare their preconditions through an \texttt{available when} clause and their effects as patches; an action is legal at turn $t$ iff its precondition holds in the current state. The DSL is non-Turing-complete by construction---computed properties are total functions, actions are first-order with bounded patches, and there is no general loop or recursion. This restriction is what makes the layer ablation well-posed: toggling \texttt{revisionEnabled} cannot accidentally alter belief update or planning logic, because each layer's actions and computed properties are syntactically scoped to its own object. Hypothetical-transition evaluation is exposed as a \texttt{sim.next} primitive that scores a candidate action by its expected one-step posterior collapse rather than by a prompt-returned claim. The runtime itself is not the contribution of this paper; it is the substrate that makes per-layer marginals measurable on a shared backend, rather than confounded by a shared prompt context. Figure~\ref{fig:loop} shows the main loop; each phase corresponds to an externally recordable set of runtime events.

\paragraph{Four harnesses.}
All four harnesses share the 500-particle Metropolis--Hastings posterior backend; they differ only in what is added on top:
\begin{itemize}\itemsep 0.1em
  \item \emph{Belief-only.} Fires at the highest-probability cell; asks no questions.
  \item \emph{+ Planning.} Evaluates shoot and question actions via \texttt{sim.next} with a two-bucket question budget.
  \item \emph{+ Symbolic reflection.} Adds an EMA-based confidence signal, a sustained-low-confidence gate, and a preset library of LLM-free revision actions (\texttt{coarse\_roi\_collapse}, \texttt{cluster\_closeout\_bias}, \texttt{late\_diffuse\_reprobe}, \ldots).
  \item \emph{+ LLM-backed revision.} Uses the same gate but may delegate the revision patch to a locally-served LLM; on protocol failure the runtime falls back to the symbolic preset.
\end{itemize}
Because belief-only and +Planning share the same belief backend, their contrast isolates the planning layer alone.

\paragraph{Setup.}
All experiments use 8$\times$8 boards with 14 ship cells, 40 shots, 15 questions, and $\varepsilon=0.1$ observation noise. We report 18 boards $\times$ 3 seeds = 54 games. Reflective defaults are: EMA coefficient $\alpha=0.25$, sustained low-confidence streak threshold 2, revision cooldown of 3 turns, minimum revision delta 0.01, and confidence threshold $\tau=0.72$ by default (swept at $\{0.0, 1.0\}$ as endpoints). The LLM is locally served \texttt{gemma4:e4b} via Ollama. The board suite is a reimplementation, not the published one of~\cite{grand2025battleship}; the cross-suite comparison is isolated to \S5 with the suite difference made explicit, rather than absorbed into the unified table.

\section{How Far the Harness Gets Us}

\begin{table}[t]
\centering
\scriptsize
\caption{Unified within-suite comparison ($54$ games = $18$ boards $\times$ $3$ seeds; $8\times8$ noisy Battleship, $\varepsilon{=}0.1$; same MCMC belief backend). \emph{LM-only Captains} call the LLM on every turn with no posterior, EIG, or decision rule; \texttt{openai/gpt-5-*} models use \texttt{reasoning\_effort=medium} (default). \emph{Harness layers} (this paper) progressively add structure on top of a shared belief; the LLM is gated to L4 only (\texttt{gemma4:e4b} via Ollama). Win-rate CI is Wilson $95\%$ on the game-completion proportion; we rely on its width to regulate the strength of claims. \textbf{Bold}: best per block.}
\label{tab:unified}
\resizebox{\columnwidth}{!}{%
\begin{tabular}{@{}lccccc@{}}
\toprule
Captain / Harness & LLM Rate & Avg F1 & Win Rate & 95\% CI & Avg Q \\
\midrule
\multicolumn{6}{@{}l}{\emph{LM-only Captains (no posterior / EIG / decision rule):}} \\
\quad gemma3n:e4b ($\sim$4B, no reasoning)            & 100\% & 0.263 & 0.0\%           & [0.0, 6.6]    & 14.2 \\
\quad Llama-4-Scout (109B MoE, no reasoning)          & 100\% & 0.353 & 0.0\%           & [0.0, 6.6]    & 14.8 \\
\quad gpt-5-nano (small reasoning)                    & 100\% & 0.459 & 29.6\%          & [19.1, 42.8]  & 12.4 \\
\quad gpt-5-mini (mid reasoning)                      & 100\% & \textbf{0.565} & \textbf{77.8\%} & [65.1, 86.8] & 14.2 \\
\midrule
\multicolumn{6}{@{}l}{\emph{Harness decomposition (this paper):}} \\
\quad L1: Belief-only                                 & 0\%   & 0.522 & 50.0\%          & [37.1, 62.9]  & 0.0 \\
\quad L2: + Planning                                  & 0\%   & 0.539 & \textbf{74.1\%} & [61.1, 83.9]  & 11.9 \\
\quad L3: + Symbolic reflection (off)                 & 0\%   & 0.552 & 57.4\%          & [44.2, 69.7]  & 8.0 \\
\quad L3: + Symbolic reflection (on)                  & 0\%   & 0.551 & 55.6\%          & [42.4, 68.0]  & 8.0 \\
\quad L4: + LLM-backed revision ($\tau{=}1.0$)        & 4.3\% & \textbf{0.557} & 53.7\% & [40.6, 66.3]  & 8.9 \\
\bottomrule
\end{tabular}%
}
\end{table}

\paragraph{Reading the unified table.}
Two facts carry the rest of this section. First, \emph{non-reasoning LM-only Captains, regardless of model size up to a $109$B MoE, fall below our no-LLM L1 baseline}: a $\sim$$4$B open-weights model lands at $0.263$ F1 (below uniform Random; \citealp{grand2025battleship} report Random F1 $0.317$ on a similar 18-board suite), Llama-4-Scout at $0.353$ ($0$ wins). Model size is not the missing ingredient---inference-time reasoning is. Second, \emph{a small reasoning model (gpt-5-nano) closes most of the L1 gap but does not cross it} ($29.6\%$ WR), while \emph{a mid-reasoning model (gpt-5-mini) reaches the L2 planning neighborhood} ($77.8\%$ WR vs.\ L2's $74.1\%$, with overlapping Wilson CIs). The same in-domain competence is therefore recoverable two ways: by adding planning structure with $0\%$ LLM calls (L1$\rightarrow$L2), or by spending an LLM call every turn at mid-reasoning class. Our decomposition isolates the no-LLM path and asks where the LLM still earns a residual \emph{within} that path; the L4 row answers quantitatively---$4.3\%$ LLM rate, $+0.005$ F1 over L3---which we read as a small residual. Read together, the unified table shifts the question from ``LLM vs.\ harness'' to harness engineering: what does each declared layer buy, and how thinly is the LLM needed once those layers are in place?

\paragraph{\S4.1 The planning layer does the heavy lifting.}
\emph{Key finding.} Adding declarative planning on top of a shared belief backend raises win rate by $+24.1$pp (50.0\% $\rightarrow$ 74.1\%) with zero LLM calls---the single largest layer effect in the decomposition, and the only contrast in which the Wilson 95\% CIs of the two harnesses ($[37.1,62.9]$ vs.\ $[61.1,83.9]$) are non-overlapping at $n{=}54$. Under the pre-specified criterion (\S1), the planning layer is therefore the unique heavy-lifting layer in this decomposition.

\emph{Mechanism.} The +24.1pp gap is not a generic ``more compute'' effect. L1 commits each turn to the cell of maximum posterior mass, which is the locally-greedy action when the posterior is sharp but offers no information value when the posterior is flat or multimodal. L2 changes the decision rule in two coupled ways. First, each candidate is scored through a \texttt{sim.next} preview that ranks actions by \emph{expected one-step posterior collapse} rather than immediate hit probability. Second, the action set includes region-DSL questions drawn under a two-bucket budget so that questions are \emph{timed}---early-game ROI narrowing and late-game cluster closeout---rather than uniformly rate-limited. The asymmetric gain pattern ($+0.017$ F1 vs.\ $+24.1$pp win rate) is what this mechanism predicts: L2 does not change local targeting precision much, but it changes how often the agent escapes the flat-posterior regime in which L1's greedy rule has no edge. The unique non-overlapping CI in the decomposition is the shape of that effect at the suite level, rather than a small mean shift.

\begin{figure}[t]
\centering
\includegraphics[width=\columnwidth]{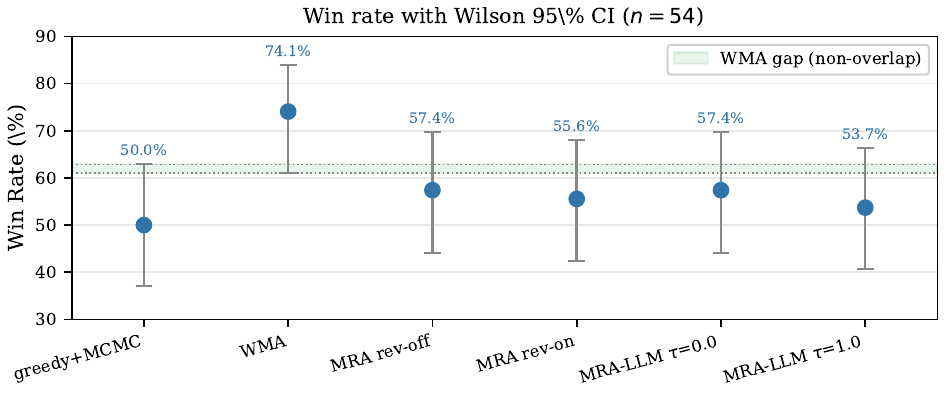}
\caption{Win rate per harness layer with Wilson $95\%$ binomial confidence interval ($n{=}54$; harness block of Table~\ref{tab:unified}). The L1$\rightarrow$L2 step is the only vertical jump in the harness; within the reflective family (L3, L4) the intervals are flat against each other and the LLM-backed L4 sits in the same band as the LLM-free L3 baselines. The picture is what ``the harness does the heavy lifting, not the LLM'' looks like at the level of confidence intervals.}
\label{fig:winrate-ci}
\end{figure}

Figure~\ref{fig:winrate-ci} renders this concentration as a single picture. The L1$\rightarrow$L2 step is the only vertical lift in the decomposition; once one moves into the reflective family the win-rate intervals are flat against each other, with the LLM-backed layer (L4) overlapping the LLM-free L3 baselines. Adding an LLM revision layer on top of an already-reflective harness produces parallel motion, not a second jump---and ``the harness does the heavy lifting, not the LLM'' is exactly the shape that picture has. The F1 gain at L1$\rightarrow$L2 is correspondingly smaller ($+0.017$), producing a \emph{last-mile asymmetry}: planning mostly converts borderline games into wins rather than changing local targeting precision. One might object that +Planning simply benefits from asking more questions. The objection misses what the planning layer does: each question is selected and timed by \texttt{sim.next} under declared budget constraints, not issued by a generic question policy.

A second, stronger objection conflates ``harness-first'' with ``LLM-free'' and reads the $+24.1$pp lift as ``you just built a good solver.'' The unified table (Table~\ref{tab:unified}) refines this within our own suite. \emph{Non-reasoning} LM-only Captains, including a $109$B MoE (Llama-4-Scout: $0.353$ F1, $0$ wins), all fall below L1 ($0.522$ F1, $50.0\%$ WR); a \emph{small reasoning} model (gpt-5-nano) closes most of the gap but does not cross it ($29.6\%$ WR); a \emph{mid-reasoning} model (gpt-5-mini) reaches the L2 planning neighborhood at $100\%$ LLM cost ($77.8\%$ WR, CI overlapping L2's). The harness's contribution is therefore not ``you don't need an LLM''---it is more pointed: the same in-domain competence is recoverable at $0\%$ LLM cost by adding $\sim$$0.017$ F1 / $+24.1$pp WR of explicit planning structure on top of a shared belief, which is the substance of the L1$\rightarrow$L2 contrast. Our decomposition then asks where the LLM still earns a residual \emph{within} that no-LLM path: at L4 the gate fires on $4.3\%$ of turns and adds $+0.005$ F1 over L3, a small qualitative residual at $n{=}54$. Read together, the unified table reframes the question from ``LLM vs.\ harness'' to harness engineering: what does each declared layer buy, and how thinly is the LLM needed once those layers are in place?

\paragraph{\S4.2 Symbolic reflection is mechanistically real but calibration-sensitive.}
\emph{Key finding.} Symbolic reflection produces large signed board-level effects (up to $\pm 0.140$ F1) that cancel on aggregate ($-0.001$ F1, $-1.8$pp win rate). The cancellation is itself a diagnosable calibration finding---two regimes firing in opposite directions---rather than absence of mechanism.

What is declared is a runtime legality condition (Listing~\ref{lst:gate}), not a prompt heuristic: revision actions are only available when low confidence is both sustained and beaten by a concrete counterfactual preview. Board-level effects are therefore large and signed in both directions:

\begin{center}\small
\begin{tabular}{@{}lc@{\qquad}lc@{}}
\toprule
\multicolumn{2}{c}{\textbf{Recovery} $\Delta$F1} & \multicolumn{2}{c}{\textbf{Over-revision} $\Delta$F1} \\
\midrule
B02 & $+0.140$ & B01 & $-0.148$ \\
B14 & $+0.099$ & B15 & $-0.085$ \\
B17 & $+0.092$ & B11 & $-0.079$ \\
B09 & $+0.076$ &     &          \\
\bottomrule
\end{tabular}
\end{center}

\begin{figure}[t]
\centering
\includegraphics[width=\columnwidth]{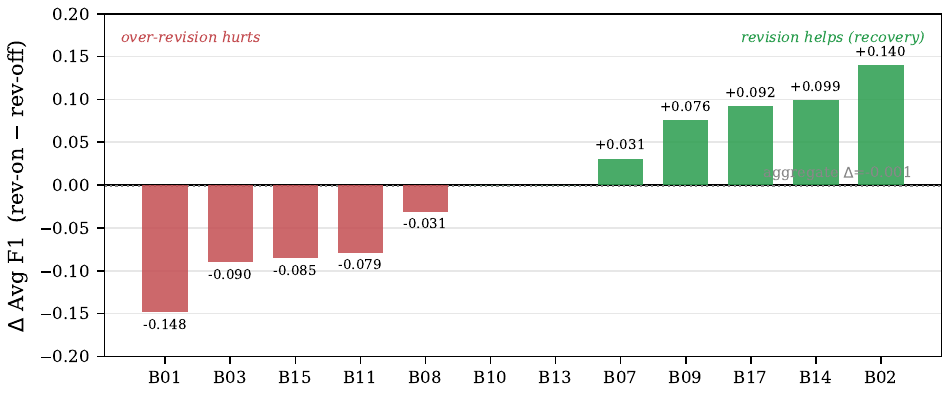}
\caption{Per-board $\Delta$F1 (symbolic reflection on minus off) across the $12$ boards recoverable from the registry top/bottom-5 overlap, sorted. Green bars are recovery boards (revision helps); red are over-revision boards (revision hurts). The dotted grey line is the aggregate $\Delta{=}-0.001$. The bimodal, signed-both-ways shape is visual evidence that the right diagnosis of the flat aggregate is calibration, not absence of mechanism.}
\label{fig:revision-delta}
\end{figure}

Figure~\ref{fig:revision-delta} plots these signed effects across all $12$ recoverable boards. The distribution is sharply bimodal---large positive bars on recovery boards, large negative bars on over-revision boards, with the aggregate $\Delta{=}-0.001$ landing exactly between the two regimes. A mechanism that did nothing would produce a flat band at zero; the actual picture is two clearly separated regimes, in opposite directions, that happen to cancel on average. The reflection mechanism is firing on both kinds of board; what is missing is a trigger calibrated to fire only on the recovery kind. The B17-seed0 trace (Appendix~\ref{app:b17}) makes the mechanism concrete. With symbolic revision enabled, \texttt{coarse\_roi\_collapse} fires at turn 2 and \texttt{cluster\_closeout\_bias} fires at turns 12 and 15, yielding a win at F1 $0.609$. With revision disabled, no revision action fires and the agent exhausts all 40 shots for F1 $0.333$. Declaring reflection as a runtime layer rather than a latent prompt pattern is what makes board-level diagnosis like this possible: the mechanism exists; what is missing is a calibrated preset library.

\paragraph{Beyond a single board.}
The B17 trace is not an isolated anecdote. Across the four recovery boards (B02, B14, B17, B09) the revision actions that fire are either \texttt{coarse\_roi\_collapse} at an early turn ($t\leq 4$) or \texttt{cluster\_closeout\_bias} after a cluster of mid-game hits---i.e., situations where the posterior genuinely spends several consecutive turns in a flat or miscalibrated regime. In the three over-revision boards (B01, B15, B11), revision instead activates on a transient low-confidence dip that resolves without intervention, diverting shots from an already-correct local targeting phase. The preset library is therefore well-calibrated for \emph{sustained} ambiguity but miscalibrated for \emph{transient} confidence dips, which is consistent with the EMA-based $c_t$ responding faster than the cooldown and streak guards can filter. This is a calibration failure of the revision trigger, not a structural failure of the reflection mechanism.

\begin{figure}[t]
\begin{lstlisting}[caption={Symbolic reflection as declared runtime legality (defaults: $\tau=0.72$, streak $\geq 2$, cooldown $=3$, $\delta_{\min}=0.01$). Confidence, sustained low-confidence, and positive preview are all computed properties rather than prompt-returned claims; \texttt{applyRevision} is a guarded action, not an LM call.},label={lst:gate}]
computed modelConfidence = 1 - (predictionErrorEMA
                               + calibrationErrorEMA) / 2
computed confident       = modelConfidence >= confidenceThreshold
computed sustained       = lowConfidenceStreak >= 2
computed canRevise       = not confident
                           and (cooldownRemaining == 0)
computed revisionRequested = canRevise and sustained
                             and positivePreview
                             and (revisionKind != "")
computed shouldRevise    = revisionEnabled and revisionRequested

action applyRevision available when shouldRevise:
    patch policyParameters <- nextParameters
    patch cooldown         <- cooldownTurns
\end{lstlisting}
\end{figure}

\paragraph{\S4.3 The LLM is a sparse, non-monotonic residual.}
\emph{Key finding.} The LLM is invoked on only $4.3\%$ of turns and its marginal effect is bounded and non-monotonic---sufficient evidence, in this setting, that the LLM is not the centre of gravity of the harness.

The confidence gate controls everything. At $\tau=0.0$ it never opens (the system collapses onto symbolic revision-off); at $\tau=1.0$ it opens on a measured $4.3\%$ of turns. The tradeoff at $\tau=1.0$ is itself interesting: F1 is the highest of the reflective family ($0.557$) while win rate is the lowest ($53.7\%$). LLM revisions sharpen local targeting at the cost of game-level completion. An 18-game vs.\ 54-game comparison (Appendix~\ref{app:sweep}) preserves the qualitative pattern (non-monotonic, bounded) but shows that the exact tradeoff is unstable under small $n$: at 18 games, $\tau=1.0$ achieves both the highest F1 \emph{and} the highest win rate; at 54 games, the win-rate advantage reverses while the F1 ranking is preserved.

\paragraph{\S4.4 What harness decomposition buys.}
Not every added layer is net-positive in aggregate---and that is the point. What decomposition buys is that mixed outcomes become \emph{interpretable} rather than averaged into an opaque pass rate:
\begin{itemize}\itemsep 0.1em
  \item Symbolic reflection: aggregate-neutral yet decisive on specific boards.
  \item LLM-backed revision: F1-helpful yet win-rate-harmful.
  \item Hidden L2$\to$L3 gap: the raw win rate also drops from $74.1\%$ (+Planning) to $57.4\%$ ($-16.7$pp) when the reflection layer is installed even with its gate off. As the Limitations section notes, this transition changes the question-budget policy as a confound, so we do \emph{not} read this as a pure reflection-layer cost---but decomposition is what surfaces it as a question to answer rather than a pattern to hide.
\end{itemize}
These patterns would not be observable in a single end-to-end pass rate. In summary, when harness layers are externalized, failure turns into diagnosable signal rather than opaque pass/fail: the primary-metric heavy-lifting layer is identified (L2), the board-level direction of L3's signed effect is made visible, L4's residual role is quantified as a dependent $4.3\%$ invocation rate, and confound-induced gaps (L2$\to$L3) are exposed rather than absorbed into an aggregate number.

\section{Discussion}

\paragraph{Harness-first, LLM-residual as a design bias.}
The decomposition supports a concrete design bias for LM-assisted planning in settings where belief, action evaluation, and revision admit symbolic representation: exhaust declarative structure first, and reserve the LLM for gated residual revision. This sharpens the direction of program-guided planners~\cite{tang2024worldcoder,liu2023llmp}: in a well-specified planning domain, the residual role of the LLM can be significantly smaller than end-to-end agent architectures suggest.

\paragraph{Measurement, not optimization, is the contribution.}
Once the harness is externalized, prediction error, model confidence, revision eligibility, and revision outcomes become first-class inspectable variables rather than latent prompt effects. Symbolic reflection is the clearest illustration: it is not yet net-positive on aggregate, yet decomposition reveals where it helps, where it hurts, and why calibration matters---as the B17-seed0 trace makes concrete. Aggregate-neutral and uninformative are different things, and the difference is only visible when the layer is lifted out.

\paragraph{Relation to metacognitive architectures.}
The decomposition connects to a long line of metacognitive systems---Soar~\cite{laird1990soar} integrates planning with impasse-driven meta-level intervention, MIDCA~\cite{cox2016midca} separates cognitive and metacognitive loops, HYDRA~\cite{piotrowski2023hydra} detects environment novelty and repairs PDDL+ domains across episodes, and self-aware agents~\cite{haber2018intrinsic} track world-model prediction error to guide exploration. Three differences matter for our setting: (i) the metacognitive loop is declared \emph{inside} the harness as guarded actions and computed signals rather than delegated to an LM wrapper or a separate cognitive cycle; (ii) revision is in-episode rather than across-episode; and (iii) the reflective loop operates without LLM calls by default, with LLM intervention attached only as a conditional effect of the same gate. This is what makes layer-by-layer ablation possible without changing the underlying mechanism.

\paragraph{F1/win-rate divergence as a budget tension.}
The F1 and win-rate signals diverge across the reflective family in a way that has a planning explanation. +Planning achieves its high win rate while asking about $11.9$ questions per game; both symbolic and LLM-backed reflective variants ask about $8$, with the LLM-backed variant recovering only to $8.9$ at $\tau{=}1.0$. Question budget is what closes out borderline games, so a revision---whether symbolic or LLM-backed---competes with question allocation for a finite per-game turn supply. LLM-chosen revisions in particular sharpen local targeting (highest F1 in the reflective family at $0.557$) but sometimes spend turns the symbolic question budget would have spent differently to finish the game. The decomposition makes this tradeoff visible at the layer level rather than absorbing it into a single end-to-end pass rate.

\paragraph{Relation to the original Battleship benchmark.}
A substantial F1 gap remains relative to the strongest published configuration of~\cite{grand2025battleship} on their suite: Llama-4-Scout combined with their full Bayesian harness (Bayes-Q + Bayes-M + Bayes-D) reaches F1 $0.764$. The plausible contributor we cannot rule out is the absence of language-informed belief construction (their LLM-emitted Python-program question pool produces richer EIG candidates than our region-based DSL)---an orthogonal LLM-in-planning axis that our decomposition does not attempt to cover. The decomposition isolates the residual role of the LLM \emph{within the planning loop}; it does not speak to the role of the LLM \emph{within the belief-construction step}, which remains an open and complementary question.

\section{Limitations}

Five caveats bound the scope of these claims.
\begin{itemize}\itemsep 0.1em
  \item \textbf{Single domain.} All results are from Battleship; the protocol is untested elsewhere. Battleship is also a probabilistic-search problem in which MCMC posteriors and \texttt{sim.next}-evaluated planning are a particularly natural fit, so the planning layer's $+24.1$pp lift may overstate what declarative planning recovers in less search-friendly domains. The same MCMC-friendly properties also stress-test the LLM's residual role from the opposite side: the unified table (Table~\ref{tab:unified}) shows LM-only Captains span F1 $0.263$--$0.565$ across non-reasoning and reasoning model classes within our suite, so the small L4 residual we measure is consistent with the symbolic substrate doing most of the planning work \emph{in this domain}, not with the LLM being unhelpful in general.
  \item \textbf{Sample size.} 54 games expose qualitative patterns but do not support tight confidence intervals; the 18-game vs.\ 54-game divergence at $\tau{=}1.0$ (Appendix~\ref{app:sweep}) is a concrete reminder of small-$n$ instability.
  \item \textbf{Preset calibration.} The current symbolic library is a first calibration pass rather than a final policy, which is why symbolic reflection is not net-positive on aggregate. Per-board signed effects suggest the mechanism is real and the gap is in calibration, not architecture.
  \item \textbf{LLM choice.} The residual was measured with one locally served 9B model (\texttt{gemma4:e4b}); the bounded, non-monotonic tradeoff may shift under larger cloud-served LLMs, and the threshold sweep is sparse rather than fully characterized.
  \item \textbf{Question-policy confound.} The transition from +Planning to +Symbolic reflection also changes the question-budget policy (two buckets to three plus exploit threshold), so the reflection ablation isolates ``reflection given its own question policy'' rather than reflection over a shared question policy. A cleaner isolation, in which the question policy is held fixed across the L2$\to$L3 transition, is left for future work.
\end{itemize}

\section{Conclusion}

Once a planning harness is externalized into measurable layers---belief tracking, declarative planning, symbolic reflection, and LLM-backed revision---the question ``how much heavy lifting can the harness do?'' becomes a layer-level decomposition rather than an opaque pass/fail. In this setting, the planning layer carries the largest single effect ($+24.1$pp win rate over a belief-only baseline, with zero LLM calls); symbolic reflection is mechanistically real but calibration-sensitive at the current preset library; and LLM-backed revision is a sparse, gated residual whose marginal effect is bounded and non-monotonic. The corresponding design bias is direct: declare what you can, reflect symbolically where possible, and reserve the LLM for the residual that the declared substrate cannot resolve. In that design, the load-bearing question for an LLM agent shifts from ``does the agent use an LLM?'' to \emph{where} LLM intervention is empirically justified, and a declared harness is the instrumentation that makes the answer measurable. Future work will (1) replicate the decomposition in a second harness domain with different action and observation geometry, (2) calibrate the symbolic preset library so reflection becomes net-positive on aggregate, and (3) characterize the full $\tau$-benefit curve across LLM sizes and quantizations.

\paragraph{Code availability.}
The declarative runtime implementation is available at \url{https://github.com/manifesto-ai/core}, and the Battleship evaluation harness---including the world models, strategy wrappers, and the \texttt{log:lens} analysis CLI used for every number in this paper---is available at \url{https://github.com/eggplantiny/battleship-manifesto}.

\appendix

\section{Threshold Sweep Detail}
\label{app:sweep}

\begin{table}[h]
\centering
\scriptsize
\caption{Threshold sweep across evaluation scales (\S4.3). Thresholds $0.0$ and $0.5$ collapse into a no-LLM basin; threshold $1.0$ activates LLM revision on a measured $4$--$5\%$ of turns. The 18-game and 54-game rows diverge at $\tau{=}1.0$, underscoring the need for larger-scale evaluation while preserving the qualitative finding of non-monotonicity.}
\label{tab:threshold-sweep}
\begin{tabular}{@{}lcccc@{}}
\toprule
Scope & $\tau$ & LLM Rate & Avg F1 & Win Rate \\
\midrule
18 games & 0.0  & 0.0\% & 0.522 & 50.0\% \\
18 games & 0.5  & 0.0\% & 0.522 & 50.0\% \\
18 games & 0.72 & 1.1\% & 0.515 & 50.0\% \\
18 games & 1.0  & 4.6\% & 0.569 & 61.1\% \\
54 games & 0.0  & 0.0\% & 0.552 & 57.4\% \\
54 games & 1.0  & 4.3\% & 0.557 & 53.7\% \\
\bottomrule
\end{tabular}
\end{table}

\section{B17-seed0 Trace Summary}
\label{app:b17}

\begin{figure}[h]
\centering
\scriptsize
\fbox{\parbox{0.92\columnwidth}{\textbf{Symbolic reflection on (win, F1 $0.609$).} Turn~2: \texttt{coarse\_roi\_collapse}. Turns~12 and 15: \texttt{cluster\_closeout\_bias}. \textbf{Symbolic reflection off (loss, F1 $0.333$).} No revision action fires; the agent exhausts 40 shots without finishing.}}
\caption{Qualitative event trace for the B17-seed0 case discussed in \S4.2. The trace is intentionally compact because the full runtime log is best inspected as an artifact bundle.}
\label{fig:b17-trace}
\end{figure}

\bibliographystyle{ACM-Reference-Format}
\bibliography{1_agent_skills_refs}

\end{document}